\documentclass[11pt]{article}

\usepackage[final]{acl}

\usepackage{times}
\usepackage{latexsym}
\usepackage[T1]{fontenc}
\usepackage[utf8]{inputenc}
\usepackage{microtype}
\usepackage{inconsolata}
\usepackage{graphicx}
\usepackage{tcolorbox}
\tcbuselibrary{skins, breakable}
\usepackage{booktabs}
\usepackage{multirow}
\usepackage{adjustbox}
\usepackage{amsmath, amsfonts}
\usepackage{xcolor}

\usepackage{amsmath}
\usepackage{mathtools}
\usepackage{amsthm}
\usepackage{multirow}
\usepackage{float}
\newtheorem{theorem}{Theorem}

\newtheorem{corollary}{Corollary}[theorem]

\usepackage{caption}
\usepackage{subcaption}
\usepackage{color, colortbl}
\definecolor{Gray}{gray}{0.9}
\newcolumntype{a}{>{\columncolor{Gray}}c}
\usepackage{wrapfig}
\usepackage{lipsum}
\usepackage{makecell}
\usepackage{algorithm}
\usepackage{algorithmic}

\def\eqref#1{Eq.~(\ref{#1})}

\title{Adaptive Guidance for Retrieval-Augmented Masked Diffusion Models}

\author{Jaemin Kim \quad \\
  Graduate School of AI, KAIST \\ Republic of Korea \\
  \texttt{kjm981995@kaist.ac.kr} \\\And
  Jong Chul Ye \\
  Graduate School of AI, KAIST \\ Republic of Korea \\
  \texttt{jong.ye@kaist.ac.kr} \\}

\begin{document}
\maketitle

\begin{abstract}
Retrieval-Augmented Generation (RAG) improves factual grounding by incorporating external knowledge into language model generation. However, when retrieved context is noisy, unreliable, or inconsistent with the model's parametric knowledge, it introduces retrieval-prior conflicts that can degrade generation quality. While this problem has been studied in autoregressive language models, it remains largely unexplored in diffusion-based language models, where the iterative denoising process introduces unique challenges for integrating retrieved context. In this work, we propose \textbf{Adaptive Retrieval-Augmented Masked Diffusion (ARAM)}, a training-free adaptive guidance framework for Masked Diffusion Models (MDMs) in RAG settings. ARAM dynamically calibrates the guidance scale during denoising according to the Signal-to-Noise Ratio (SNR) of the distributional shift induced by retrieved context. Intuitively, the model strengthens guidance when the retrieved context provides reliable corrective evidence and suppresses it when the contextual signal is noisy or non-supportive. Extensive experiments on multiple knowledge-intensive QA benchmarks show that ARAM improves overall QA performance over competitive RAG baselines.

\end{abstract}

\section{Introduction}

Large Language Models (LLMs) have demonstrated remarkable capabilities across various natural language processing tasks~\cite{grattafiori2024llama, hui2024qwen2}.  However, their reliance on static parametric knowledge often leads to hallucinations and outdated responses when dealing with dynamic or knowledge-intensive queries. Retrieval-Augmented Generation (RAG)~\cite{lewis2020retrieval, zhao2026retrieval} addresses this limitation by grounding generation in external knowledge retrieved at inference time. 

Recently, diffusion-based language generation has emerged as a promising alternative to autoregressive models. In particular, Masked Diffusion Models (MDMs) generate text through an iterative denoising process that progressively fills masked tokens while leveraging bidirectional attention~\cite{llada, dream}. Compared to autoregressive models, this generation paradigm enables parallel token updates and iterative refinement of intermediate predictions.

\begin{figure}[t]
    \centering
    \includegraphics[width=\linewidth]{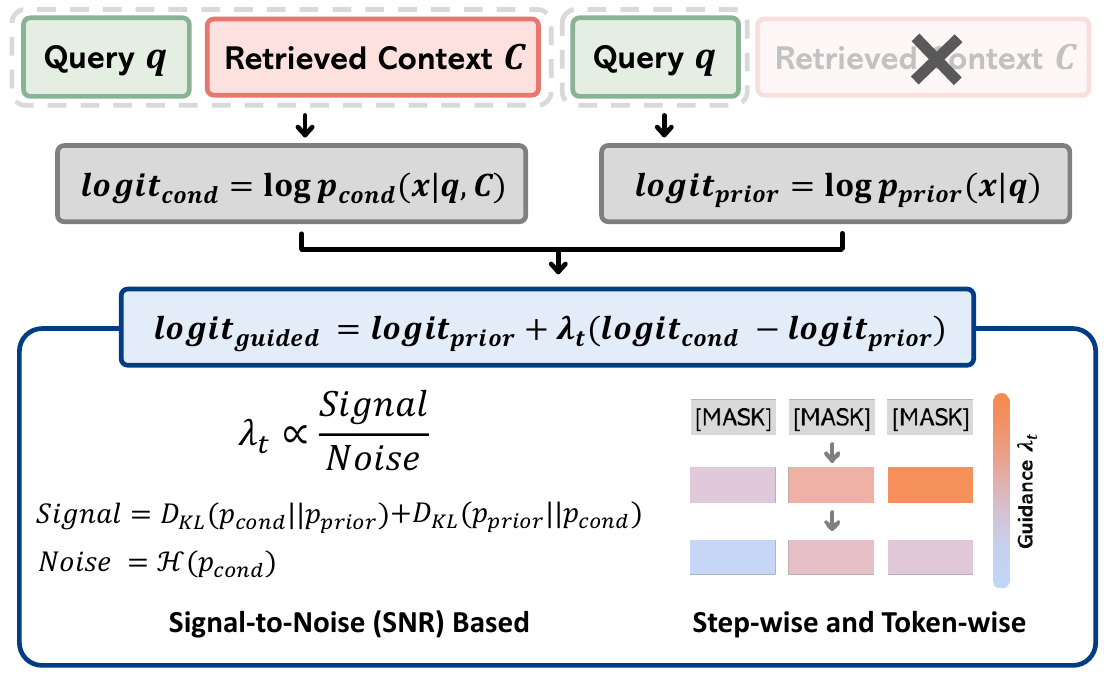}
    \vspace{-0.7cm}
\caption{\textbf{Overview of ARAM.} ARAM mitigates knowledge conflict via step-wise and token-wise adaptive guidance driven by the Signal-to-Noise Ratio.}
    \label{fig:main}
    \vspace{-0.4cm}
\end{figure}

While recent research has focused on improving the generative capabilities of MDMs to narrow the gap with autoregressive models~\cite{lou2023discrete}, applying RAG to MDMs remains notably underexplored. Preliminary studies indicate that MDM-based RAG demonstrates remarkable in-context learning capabilities that often surpass those of autoregressive models~\cite{dai2026revealing, yu2026unlocking}. Despite this potential, existing approaches predominantly assume that the retrieved documents are reliable. In realistic RAG settings, however, the retrieved context is often noisy, partially irrelevant, or even contradictory to the model's internal knowledge.

A fundamental challenge is how to effectively ground the iterative denoising process in retrieved knowledge, especially when this external information conflicts with the model's parametric prior. When such heterogeneous context is injected into the generation process, the model must dynamically decide how much it should trust the retrieved context versus its parametric prior. We refer to this tension between retrieved context and parametric knowledge as {\em knowledge conflict}~\cite{shi2024trusting, wang2025retrieval, yuan2024discerning}.
This raises a key question:
\begin{figure}[!h]
    \centering
    \begin{tcolorbox}[colback=gray!5!white, colframe=gray!75!black, fonttitle=\bfseries]
    How should a diffusion language model determine when to trust retrieved context and when to rely on its parametric knowledge?
    \end{tcolorbox}
    \vspace{-0.4cm}
\end{figure}

To address this question, we propose \textbf{Adaptive Retrieval-Augmented Masked Diffusion (ARAM)}, an adaptive guidance framework that self-calibrates the guidance scale at each denoising step and token position. To effectively inject conditional information and control the generation quality in MDMs, we adopt Classifier-Free Guidance (CFG)~\cite{cfg} for RAG. However, applying a static guidance scale under retrieved contexts leads to a critical dilemma: (1) over-guidance on noisy context, which causes hallucinations, and (2) under-guidance on valid context, which fails to correct the flawed prior.

Our key insight is that the influence of retrieved context can be interpreted as a distributional shift from a retrieval-free prior to a retrieval-conditioned posterior. Specifically, 
we analyze the interaction through an information-theoretic lens, and derive a  Signal-to-Noise Ratio (SNR) of this distributional shift to adapt the guidance strength. 
Consequently, when retrieved context consistently supports the conditional distribution, guidance can be strengthened; when the retrieved context is uncertain or non-supportive, guidance can be attenuated to prevent the over-amplification of noise. We illustrate the overview of our proposed method in Fig.~\ref{fig:main}.

We evaluate our adaptive guidance mechanism on  knowledge-intensive QA benchmarks. Empirical results demonstrate that our method outperforms both static CFG and baseline MDMs, achieving significant improvements in exact matching (EM) and F1 scores for MDM-based RAG. Our main contributions are as follows:
\begin{itemize}
    \item We propose ARAM, an adaptive CFG framework for MDMs in RAG, which dynamically modulates guidance scale based on the SNR.
    
    \item We analyze the interaction between retrieval context and parametric prior in diffusion-based generation through an information-theoretic lens and derive a guidance rule motivated by signal and uncertainty.
    
    \item Extensive experiments across multiple QA benchmarks validate that our method improves EM and F1 over other baselines and mitigates retrieval-prior conflicts.
\end{itemize}

\section{Related Work}

\subsection{Knowledge Conflict in RAG}

While Retrieval-Augmented Generation (RAG) improves factual grounding by incorporating external context, it can also introduce knowledge conflict~\cite{longpre2021entity} between the model's internal knowledge and the retrieved external context, leading to degraded performance and hallucinations~\cite{cheng2024understanding}. To mitigate these conflicts, current approaches rely on external context assessors~\cite{choi2025conflict}, multi-agent debates~\cite{wang2025retrieval}, or contrastive decoding~\cite{shi2024trusting, wang2025adacad}. Among these, CAD~\cite{shi2024trusting} is most closely related to our work, as it contrasts output distributions with and without external context to penalize reliance on parametric priors. However, applying a static contrastive penalty often overcorrects when the retrieved context already aligns with the model's parametric knowledge, inadvertently degrading performance on non-conflicting data~\cite{yuan2024discerning}. To address this overcorrection, dynamic decoding strategies like AdaCAD~\cite{wang2025adacad} and COIECD~\cite{yuan2024discerning} have emerged. While effective, these approaches are fundamentally designed for autoregressive language models and cannot be trivially extended to the non-causal, iterative denoising dynamics of diffusion-based language models. Motivated by the need for dynamic adjustment~\cite{wang2025adacad, yuan2024discerning}, we introduce a discrete-diffusion-specific framework that modulates the guidance scale at each denoising step using a Signal-to-Noise Ratio (SNR).

\subsection{Classifier-Free Guidance in MDMs}
Classifier-Free Guidance (CFG)~\cite{cfg} is widely used to strengthen conditional generation by combining conditional and prior score functions. Since CFG was originally formulated for continuous spaces, defining an analogous procedure for discrete diffusion is nontrivial, as the discrete state space is inherently non-differentiable.
Recent theoretical works~\cite{schiff2024simple, rojas2025theory} have formalized the validity of CFG at the logit level in discrete diffusion models, but standard CFG applies a static guidance scale throughout the entire denoising process. Recognizing the dynamic uncertainty during generation, recent methods like A-CFG~\cite{li2025adaptive} dynamically alter the unconditional input by re-masking low-confidence tokens. Distinct from modifying the input formulation, our approach directly calibrates the guidance itself, preventing over-guidance on noisy context and under-guidance on valid corrections.

\subsection{RAG in Masked Diffusion Models}

Recent studies demonstrate the potential of integrating RAG with MDMs to enhance knowledge utilization beyond autoregressive (AR) models. 
For example, recent analysis on Attention Floating~\cite{dai2026revealing} reveals that MDMs effectively capture semantically relevant information and mitigate the influence of distracting documents, leveraging the dynamic nature of attention sinks during the denoising process~\cite{rulli2025attention}.
This behavior allows MDMs to effectively track relevant context, resulting in stronger utilization of external knowledge than AR models. SPREAD~\cite{yu2026unlocking} further shows that MDMs exhibit a strong reliance on contextual information and improve answer precision by enforcing query-level semantic coherence. Despite this potential, integrating RAG into MDMs remains unexplored. To the best of our knowledge, our work is the first to address knowledge conflict in MDM-based RAG through an intrinsic step-wise adaptive guidance mechanism.

\section{Preliminaries}

\paragraph{Masked Diffusion Language Models.}
Masked Diffusion Models (MDMs) generate text through an iterative denoising process over discrete token sequences. Let $\mathbf{x}_0 = (x_1, \dots, x_L)$ denote the target token sequence of length $L$, where each token $x_i$ belongs to a vocabulary $\mathcal{V}$. The forward process $q(\mathbf{x}_t | \mathbf{x}_{t-1})$ gradually corrupts the sequence over discrete timesteps $t \in \{1, \dots, T\}$ by replacing a subset of tokens with a special mask token $\texttt{[MASK]}$. The reverse process is parameterized by a neural network $p_\theta$, which predicts token distributions $p_\theta(\mathbf{x}_0 | \mathbf{x}_t)$ for the masked positions. The state transition to the previous step $\mathbf{x}_{t-1}$ is then sampled using the posterior distribution:
\begin{equation}
    p_\theta(\mathbf{x}_{t-1} | \mathbf{x}_t) = \sum_{\mathbf{x}_0} q(\mathbf{x}_{t-1} | \mathbf{x}_t, \mathbf{x}_0) p_\theta(\mathbf{x}_0 | \mathbf{x}_t).
\end{equation}
In MDMs, this unmasking process is typically executed in parallel across multiple masked positions based on token-level confidence. For notational simplicity, the following definitions are written for a single masked position, where $x_t$ denotes the noised token at timestep $t$ and $x$ denotes the token at the target position.

\paragraph{Retrieval-Augmented Generation.}
In Retrieval-Augmented Generation (RAG), the model conditions generation on a user query $q$ and a set of $K$ retrieved documents $\mathcal{C} = \{d_1, \dots, d_K\}$. The retrieved context $\mathcal{C}$ is concatenated with the query and provided as input to the language model. We denote this conditional token distribution as:
\begin{equation}
    p_{\text{cond}}(x):= p_{\theta}(x | x_t, q, \mathcal{C}).
\end{equation}
Conversely, when the model relies on its parametric prior without retrieval context, the distribution is:
\begin{equation}
    p_{\text{prior}}(x) := p_{\theta}(x | x_t, q).
\end{equation}

\paragraph{Classifier-Free Guidance in Discrete Diffusion.}
While Classifier-Free Guidance (CFG) was originally designed for continuous diffusion by extrapolating score differences, CFG in discrete state spaces operates directly on the prediction logits. Let $\ell_{\text{cond}}$ and $\ell_{\text{prior}}$ denote the logits corresponding to $p_{\text{cond}}$ and $p_{\text{prior}}$, respectively. The guided logits are computed as follows, where $\lambda \ge 0$ is the guidance scale:
\begin{equation}
\label{eq:static_cfg}
    \ell_{\lambda} = \ell_{\text{prior}} + \lambda (\ell_{\text{cond}} - \ell_{\text{prior}}).
\end{equation}

\section{Method}

In this section, we present the Adaptive Retrieval-Augmented Masked Diffusion (ARAM) framework. We first mathematically define the knowledge conflict and the limitations of static guidance (Sec.~\ref{sec:knowledge_conflict}), and 
then interpret CFG as an exponential tilting family (Sec.~\ref{subsec:cfg_tilting}).  Finally, we derive an adaptive guidance rule based on an information-theoretic Signal-to-Noise Ratio (SNR) (Sec.~\ref{subsec:dv_snr} and Sec.~\ref{subsec:snr_proxy}).

\subsection{Dilemma of Static Guidance}
\label{sec:knowledge_conflict}

In standard discrete CFG (\eqref{eq:static_cfg}), the guidance scale $\lambda$ uniformly amplifies the logit difference across all tokens and diffusion steps. However, in realistic RAG settings, the retrieved context $\mathcal{C}$ introduces varying degrees of reliability, creating a knowledge conflict between the parametric prior and external evidence~\cite{longpre2021entity}. 
We can categorize these generation dynamics into three distinct scenarios:

\begin{enumerate}
    \item \textbf{Reliable Context:} When $\mathcal{C}$ contains highly relevant and accurate supporting facts, $p_{\text{cond}}$ confidently shifts toward the correct tokens. This produces a large divergence between $p_{\text{cond}}$ and $p_{\text{prior}}$ with low uncertainty. In this scenario, a larger $\lambda$ is required to fully incorporate the retrieved facts and update missing or outdated parametric knowledge.
    
    \item \textbf{Irrelevant Context:} When $\mathcal{C}$ is topically unrelated to the query, the model largely ignores the external context, resulting in minimal distributional divergence  ($p_{\text{cond}} \approx p_{\text{prior}}$). The guidance scale $\lambda$ should remain close to zero to prevent unnecessary perturbation of the generation process.
    
    \item \textbf{Conflicting Context:} When $\mathcal{C}$ contains misleading, unreliable, or  contradictory information relative to the query or the model prior, $p_{\text{cond}}$ can deviate substantially from $p_{\text{prior}}$. However, this deviation is typically accompanied by high uncertainty or internal inconsistency. If a large $\lambda$ is applied, it overwrites the parametric prior $p_{\text{prior}}$ and exacerbates hallucinations. Therefore, $\lambda$ must be adaptively suppressed considering the uncertainty of $p_{\text{cond}}$.
\end{enumerate}

To resolve these varying dynamics, the model requires an intrinsic, token-level mechanism to evaluate the trustworthiness of the condition $\mathcal{C}$ and adaptively modulate $\lambda$.

\subsection{Information-Theoretic View of CFG}
\label{subsec:cfg_tilting}
We consider an MDM generating a discrete token sequence. To quantify the influence of the retrieved context $\mathcal{C}$ at diffusion step $t$, we define the \emph{context score} as a log-likelihood ratio:
\begin{equation}
    s(x;\mathcal{C}) := \log \frac{p_{\text{cond}}(x)}{p_{\text{prior}}(x)}.
\end{equation}
Using this term with \eqref{eq:static_cfg}, the guided distribution $p_\lambda(x) \propto  p_{\text{cond}}(x)^\lambda p_{\text{prior}}(x)^{1-\lambda}$ can then be written as:
\begin{equation}
    p_\lambda(x) = \frac{p_{\text{prior}}(x)\exp(\lambda s(x;\mathcal{C}))}{Z_\lambda},
\end{equation}
where $Z_\lambda = \sum_{x \in \mathcal{V}} p_{\text{prior}}(x)\exp(\lambda s(x;\mathcal{C})).$

To quantify the influence of retrieved context, we refer to the retrieval information gain as:
\begin{equation}
    \mathrm{IG}_t := D_{\mathrm{KL}}(p_{\text{cond}} \| p_{\text{prior}}) = \mathbb{E}_{p_{\text{cond}}}[s(x;\mathcal{C})].
\end{equation}
This quantity measures how strongly the retrieved context alters the model's posterior belief. If the retrieved context $\mathcal{C}$ is treated as a random variable produced by the retriever, the Conditional Mutual Information (CMI) satisfies
\begin{equation}
I(x_0;\mathcal{C}\ | x_t)
= \mathbb{E}_{\mathcal{C}}[D_{\mathrm{KL}}(p_{\text{cond}}\|p_{\text{prior}})]
= \mathbb{E}_{\mathcal{C}}[\mathrm{IG}_t].
\end{equation}
Thus, $\mathrm{IG}_t$ serves as an information-theoretic signal measuring the usefulness of retrieved context.

\subsection{Lower Bound and the Ideal SNR}
\label{subsec:dv_snr}
To understand the optimal choice of $\lambda$, we relate the information gain to the Donsker-Varadhan (DV) variational representation~\cite{donsker1983asymptotic}.
All proofs are provided in Appendix~\ref{sec:proof}.

\begin{theorem}[DV Variational Lower Bound]
\label{theorem:1}
For any $\lambda$, the retrieval information gain satisfies:
\begin{equation}
    \mathrm{IG}_t \ge \mathcal{L}(\lambda) := \lambda \mathbb{E}_{p_{\text{cond}}}[s(x;\mathcal{C})] - \log Z_\lambda.
    \label{eq:dv_lower_bound}
\end{equation}
\end{theorem}

\begin{corollary}[Global Maximizer of the DV Bound]
\label{coro:dv_max}
Let $\mathcal{L}(\lambda)$ be defined in ~\eqref{eq:dv_lower_bound}. Then $\mathcal{L}(1)=D_{\mathrm{KL}}(p_{\text{cond}} \| p_{\text{prior}})$. Hence, $\lambda=1$ maximizes the variational lower bound $\mathcal{L}(\lambda)$.
\end{corollary}

If the retrieved context $\mathcal{C}$ were a perfect oracle, Corollary~\ref{coro:dv_max} suggests that setting $\lambda=1$ would be globally optimal. However, in practical RAG settings (Sec.~\ref{sec:knowledge_conflict}), the retriever may return noisy or conflicting documents in $\mathcal{C}$. Therefore, setting $\lambda=1$ assumes full trust in $p_{\text{cond}}$ and leaves the generation highly vulnerable to hallucinations. 

To mitigate this risk, we treat the prior distribution $p_{\text{prior}}$ (where $\lambda=0$) as a safe parametric anchor. Instead of jumping to the global maximum at $\lambda=1$, 
we analyze the local geometry around our anchor ($\lambda=0$) by taking a second-order Taylor expansion as:
\begin{equation}
    \mathcal{L}(\lambda)
    \approx
    \lambda \underbrace{\big(\mathbb{E}_{p_{\text{cond}}}[s]-\mathbb{E}_{p_{\text{prior}}}[s]\big)}_{\text{Signal}}
    - \frac{\lambda^2}{2} \underbrace{\mathrm{Var}_{p_{\text{prior}}}(s)}_{\text{Noise }}.
    \label{eq:newton_snr_ideal}
\end{equation}
using the derivatives $\mathcal{L}'(0) = \mathbb{E}_{p_{\text{cond}}}[s] - \mathbb{E}_{p_{\text{prior}}}[s]$ and $\mathcal{L}''(0) = -\mathrm{Var}_{p_{\text{prior}}}(s)$. Maximizing this concave approximation yields
\begin{equation}
\lambda^* = \frac{\text{Signal}}{\text{Noise}} = 
\frac{\mathbb{E}_{p_{\text{cond}}}[s]-\mathbb{E}_{p_{\text{prior}}}[s]}
{\mathrm{Var}_{p_{\text{prior}}}(s)} .
\label{eq:optimal_lambda}
\end{equation}
This equation naturally forms the Signal-to-Noise Ratio (SNR): an expected information gain (Signal) penalized by the dispersion introduced by the context (Noise). Detailed proofs are provided in Appendix~\ref{sec:proof}.

\subsection{Practical Adaptive Guidance}
\label{subsec:snr_proxy}

\begin{figure}[!t]
    \centering
    \begin{subfigure}[b]{0.48\columnwidth}
        \centering
        \includegraphics[width=\textwidth]{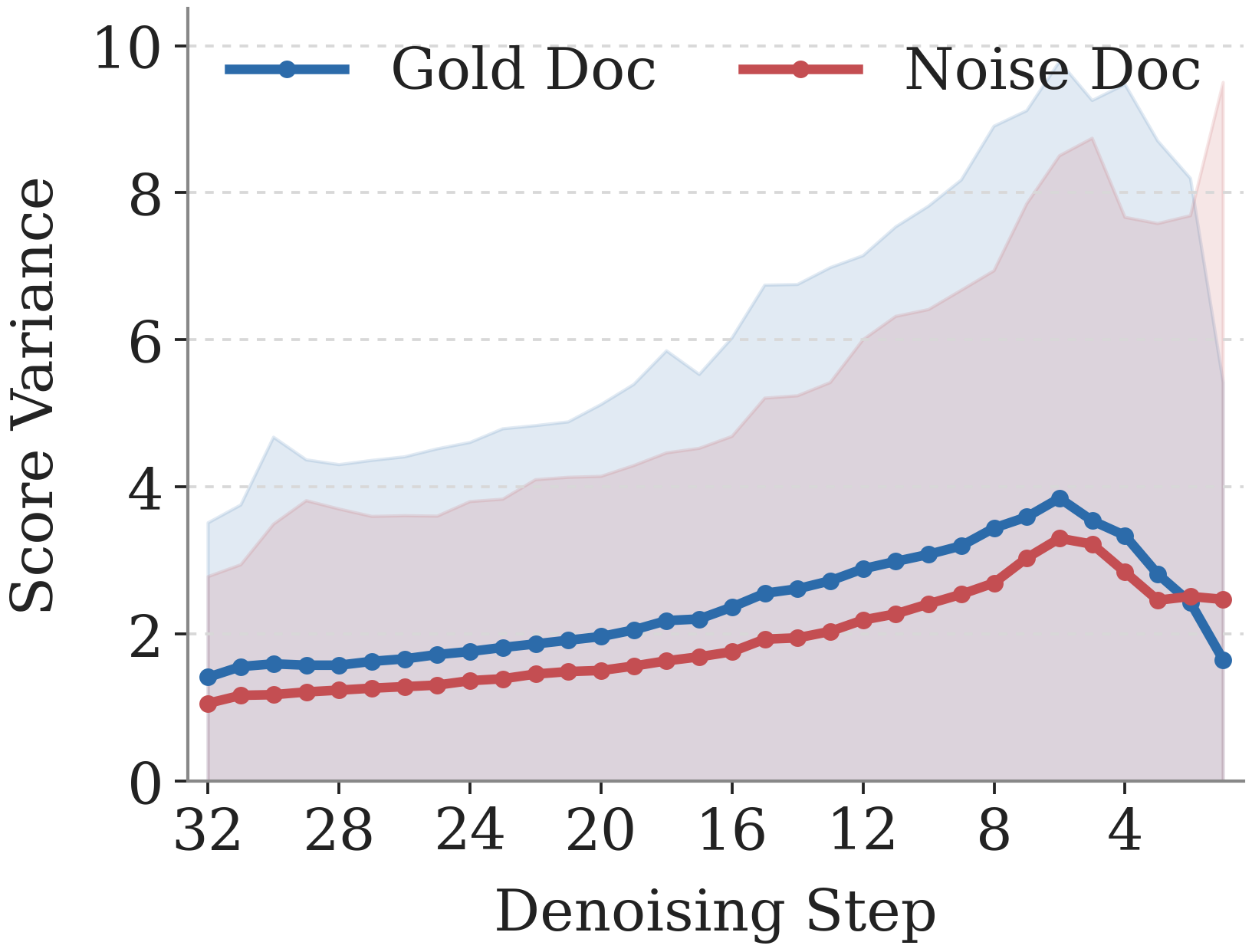}
        \caption{$\mathrm{Var}_{p_{\text{prior}}}(s)$}
        \label{fig:variance_unstable}
    \end{subfigure}
    \hfill
    \begin{subfigure}[b]{0.48\columnwidth}
        \centering
        \includegraphics[width=\textwidth]{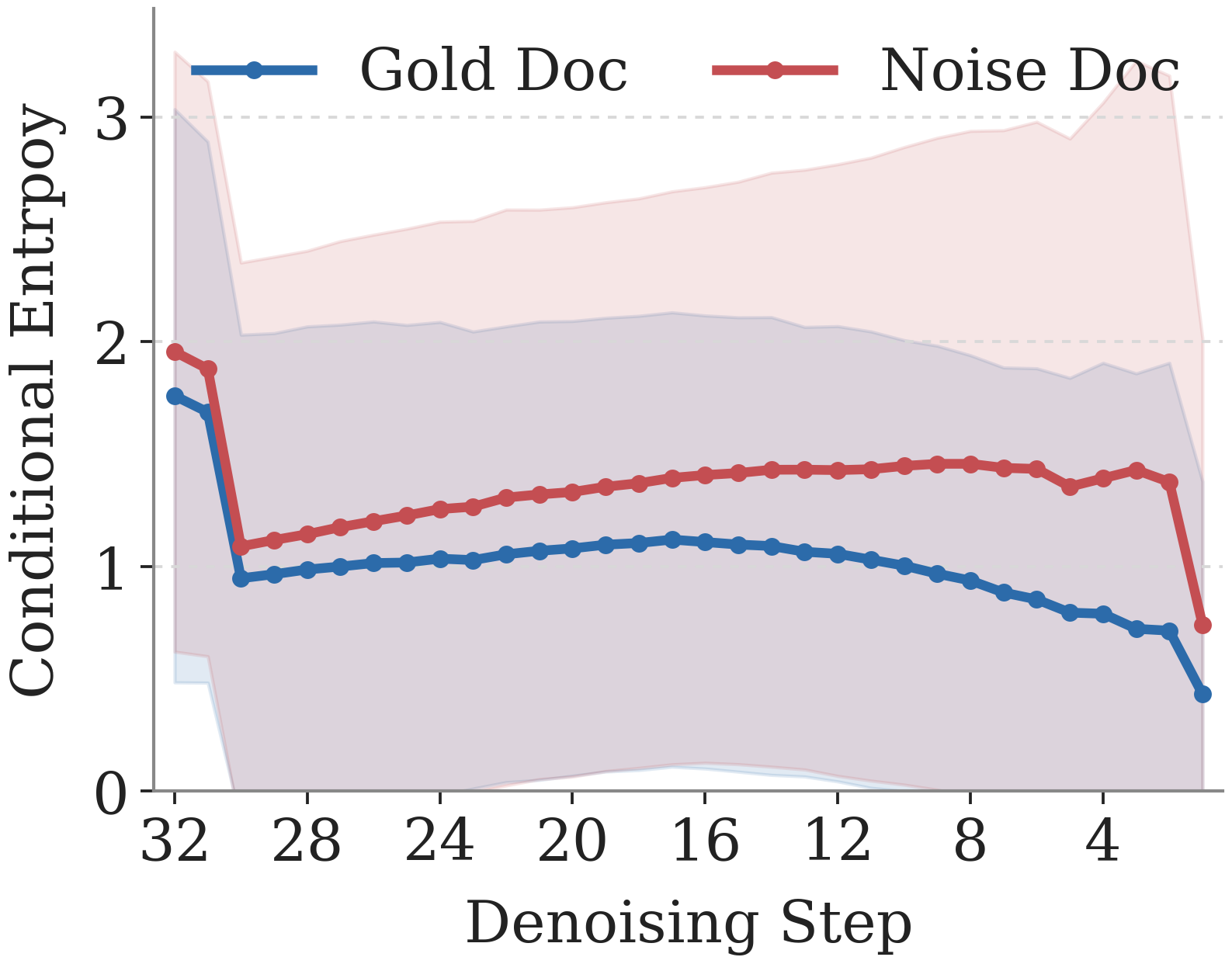}
        \caption{$\mathcal{H}(p_{\text{cond}})$}
        \label{fig:entropy_stable}
    \end{subfigure}
    \caption{\textbf{Comparison of noise proxies during the denoising process.} \textbf{(a)} The variance of the log-likelihood ratio $s$ attenuates the guidance scale even when correction is required. \textbf{(b)} In contrast, conditional entropy provides a stable measure of uncertainty, more clearly distinguishing between informative and noisy contexts.}
    \label{fig:noise_proxy_comparison}
    \vspace{-0.5cm}
\end{figure}

Although \eqref{eq:optimal_lambda} provides a theoretically optimal $\lambda^*$, directly using $\mathrm{Var}_{p_{\text{prior}}}(s)$ as the noise term is unstable in masked diffusion decoding. As shown in Fig.~\ref{fig:variance_unstable}, the variance often becomes larger for \textit{Gold Doc} (documents containing the correct answer) than for \textit{Noise Doc} (documents not containing the answer), which undesirably suppresses $\lambda_t$ even when the retrieved context is informative. In addition, the variance exhibits substantial step-wise fluctuations, making the resulting guidance scale unstable across denoising steps. Experimental details are provided in Sec.~\ref{sec:dynamics}.

To obtain a more stable estimate of the dispersion introduced by the context, we instead use the conditional entropy $\mathcal{H}(p_{\text{cond}})$ as the noise proxy. This choice is motivated by prior observations that natural language generation typically lies within a narrow entropy region, and deviations from this region indicate conflicting or uncertain contextual signals~\cite{arora2023stable, yuan2024discerning}. As shown in Fig.~\ref{fig:entropy_stable}, when the retrieved context provides clear and consistent information, the conditional distribution becomes concentrated and its entropy decreases. Conversely, when the retrieved context introduces ambiguity or contradiction, the conditional distribution becomes more diffuse, resulting in higher entropy. Accordingly, we use $\mathcal{H}(p_{\text{cond}})$ as a practical surrogate for contextual noise, which empirically preserves the desired behavior and SNR structure of the optimal guidance.

In practice, for each denoising step $t$, we compute the following signal and noise proxies:
\begin{align*}
    \mathrm{Signal} &:= \mathbb{E}_{p_{\text{cond}}}[s]-\mathbb{E}_{p_{\text{prior}}}[s] \\ &= D_{\mathrm{KL}}(p_{\text{cond}} \| p_{\text{prior}}) + D_{\mathrm{KL}}(p_{\text{prior}} \| p_{\text{cond}}), \\
    \mathrm{Noise}  &:= \mathcal{H}(p_{\text{cond}}) = -\sum_{x\in\mathcal{V}} p_{\text{cond}}(x)\log p_{\text{cond}}(x).
\end{align*}
To ensure stability and controllability, we scale the SNR:
\begin{equation}
    \lambda_t = \lambda_{\max} \cdot \tanh \left(\beta\cdot {\frac{\mathrm{Signal}}{\mathrm{Noise} + \epsilon}} \right),
    \label{eq:lambda_impl}
\end{equation}
where $\lambda_{\max}$ defines the maximum guidance scale, $\beta$ controls the sensitivity to the SNR, the $\tanh$ function and $\epsilon >0$ ensure numerical stability. Finally, we apply this token-wise adaptive $\lambda_t$ to the logit-space CFG interpolation:
\begin{equation}
\label{eq:logit_cfg_aram}
    \ell_{\mathrm{guided}}(x) = \ell_{\text{prior}}(x) + \lambda_t \big(\ell_{\text{cond}}(x)-\ell_{\text{prior}}(x)\big).
\end{equation}
This adaptive guidance is applied independently to each token at every denoising step. The complete procedure for our proposed ARAM method is detailed in Algorithm~\ref{alg:aram}.

\begin{algorithm}[t]
\caption{Adaptive Retrieval-Augmented Masked Diffusion (ARAM)}
\label{alg:aram}
\begin{algorithmic}[1]
\linespread{1.1}\selectfont 
\small
\REQUIRE Query $q$, Retrieved context $\mathcal{C}$, Masked Diffusion Model $p_\theta$, Timesteps $T$, Maximum Guidance Scale $\lambda_{\max}$, Sensitivity $\beta$, A Small Positive $\epsilon$
\STATE Initialize $\mathbf{x}_T \leftarrow (\texttt{[MASK]}, \dots, \texttt{[MASK]})$
\FOR{$t = T$ \textbf{to} $1$}
    \STATE $p_{\text{prior}}, p_{\text{cond}} \leftarrow p_\theta(\cdot | \mathbf{x}_t, q), p_\theta(\cdot | \mathbf{x}_t, q, \mathcal{C})$
    \FOR{Each masked position}
    \STATE $ \text{Signal} \leftarrow D_{\mathrm{KL}}(p_{\text{cond}}\|p_{\text{prior}})\!+\!D_{\mathrm{KL}}(p_{\text{prior}}\|p_{\text{cond}})$

    \STATE $\text{Noise} \!\leftarrow \mathcal{H}(p_{\text{cond}})$
    \STATE $\lambda_{t} \leftarrow \lambda_{\max} \cdot \tanh \left(\beta \cdot \frac{\text{Signal}}{\text{Noise} + \epsilon} \right)$
    \STATE Compute $\ell_{\text{guided}}$ using \eqref{eq:logit_cfg_aram}
    \ENDFOR
\STATE Sample $\mathbf{x}_{t-1}$ using the guided logits $\ell_{\text{guided}}$ according to the unmasking policy
\ENDFOR
\RETURN Generated sequence $\mathbf{x}_0$
\end{algorithmic}
\end{algorithm}

\begin{table*}[t!]
\centering
\small
\caption{Performance comparison across QA benchmarks.  Bold indicates the best score within each backbone.}
\label{tab:main_results}
\adjustbox{width=0.9\textwidth}{
\begin{tabular}{lcccccccccc}
\toprule
\multicolumn{1}{c}{\multirow[c]{4}{*}{\textbf{Method}}}
& \multicolumn{6}{c}{\textbf{Open-Domain QA}} 
& \multicolumn{2}{c}{\textbf{Multi-hop QA}} 
& \multicolumn{2}{c}{\textbf{Slot Filling}} 
\\
\cmidrule(lr){2-7} \cmidrule(lr){8-9} \cmidrule(lr){10-11}

& \multicolumn{2}{c}{\textbf{NQ}}
& \multicolumn{2}{c}{\textbf{TriviaQA}}
& \multicolumn{2}{c}{\textbf{MARCO QA}}
& \multicolumn{2}{c}{\textbf{HotpotQA}}
& \multicolumn{2}{c}{\textbf{T-REx}}
\\

\cmidrule(lr){2-3}\cmidrule(lr){4-5}\cmidrule(lr){6-7}\cmidrule(lr){8-9}\cmidrule(lr){10-11} 

& EM & F1 & EM & F1 & EM & F1 & EM & F1 & EM & F1 
\\
\midrule
\multicolumn{11}{c}{\textbf{Autoregressive Models}} \\

\midrule
LLaMA & 20.5 & 30.9 & 64.4 & 69.3 & 1.4 & 14.7 & 15.0 & 22.8 & 34.2 & 39.6 
\\
+ RAG~\cite{lewis2020retrieval} & 31.9 & 48.2 & 71.7 & 78.3 & 2.4 & 19.0 & 21.2 & 31.4 & 30.3 & 35.3 
\\
Qwen & 14.9 & 25.2 & 48.1 & 52.7 & 1.1 & 11.8 & 21.2 & 29.0 & 32.6 & 36.5 
\\
+ RAG~\cite{lewis2020retrieval} & 31.3 & 46.6 & 70.8 & 76.6 & 2.3 & 19.0 & 24.2 & 33.7 & 30.1 & 34.7
\\
 
\midrule
\multicolumn{11}{c}{\textbf{Masked Diffusion Models  - LLaDA}} \\
\midrule

LLaDA & 9.1 & 15.9 & 27.7 & 32.0 & 1.1 & 9.7 & 15.7 & 21.5 & 23.9 & 26.4 
\\
+ RAG~\cite{lewis2020retrieval} & 17.5 & 31.9 & 44.8 & 55.1 & 1.4 & \textbf{25.6} & 6.5 & 15.6 & 17.8 & 23.8 
\\

+ CAD~\cite{shi2024trusting} &  1.4	&16.6	&7.1	&23.9	&0.4	&24.7	&0.3	&9.0	&1.3	&7.4	
\\
+ \textsc{AdaCAD} ~\cite{wang2025adacad}& 9.3	&18.2	&27.1	&33.6	&1.2	&14.3&	15.0	&21.4	&21.7	&24.7
\\
+ COIECD~\cite{yuan2024discerning} & 13.0	&28.4	&37.4&	49.2	&0.9	&25.4	&7.1	&15.7	&11.1&	17.1	
\\
+ SPREAD~\cite{yu2026unlocking} & 19.3&	34.4	&51.7	&61.0	&1.7	&25.2&	12.6	&21.2	&18.0&	24.0	
\\
+ A-CFG~\cite{li2025adaptive} & 19.1	&33.9	&51.8	&61.2	&1.4	&24.5	&12.6	&21.5	&18.2	&\textbf{35.9}	
\\
\rowcolor{cyan!10}
+ \textbf{Ours} & \textbf{30.3} & \textbf{43.2} & \textbf{67.7} & \textbf{73.1} & \textbf{2.5} & 20.0 & \textbf{20.7} & \textbf{28.9} & \textbf{31.4} & 34.8 
\\
\midrule
\multicolumn{11}{c}{\textbf{Masked Diffusion Models  - Dream}} \\
\midrule

Dream & 14.9 & 24.9 & 36.7 & 43.5 & 1.9 & 11.0 & 16.9 & 26.2 & 30.4 & 35.1 
\\
+ RAG~\cite{lewis2020retrieval} & 35.9 & 50.5 & 72.7 & 78.4 & 2.8 & 17.5 & 25.0 & 35.2 & 29.7 & 35.0 
\\

+ CAD~\cite{shi2024trusting} & 23.9	&40.9	&55.5	&67.0	&1.7	&20.9	&14.1	&24.8	&19.3	&26.8	
\\
+ \textsc{AdaCAD}~\cite{wang2025adacad} & 37.4	&50.7	&72.5	&77.4	&2.9	&\textbf{21.7}	&24.3	&34.5	&31.4	&36.7	
\\
+ COIECD~\cite{yuan2024discerning} &30.2	&45.9	&68.5	&75.4	&2.7	&\textbf{21.7}	&19.1	&29.4	&24.4	&30.6
\\
+ SPREAD~\cite{yu2026unlocking} &35.9	&50.1	&73.7	&78.5	&2.9	&17.8	&25.0	&34.7	&\textbf{34.5}	&\textbf{38.0}
\\
+ A-CFG~\cite{li2025adaptive}& 34.0	&49.2	&70.6	&77.8	&2.6	&17.4	&24.2	&34.6	&28.1	&34.1	
\\
\rowcolor{cyan!10}
+ \textbf{Ours} & \textbf{37.7} & \textbf{51.4} & \textbf{74.4} & \textbf{79.5} & \textbf{3.3} & 18.0 & \textbf{25.8} & \textbf{36.1} & 30.0 & 35.5
\\
\bottomrule
\end{tabular}
}
\end{table*}

\section{Experiments and Results}
\subsection{Experimental Protocol}
\paragraph{Datasets and evaluation metrics.}
Following \citet{dai2026revealing}, we evaluate our method on five diverse knowledge-intensive QA datasets: NQ~\cite{nq}, TriviaQA~\cite{triviaqa}, MARCO QA~\cite{marcoqa}, HotpotQA~\cite{hotpotqa}, and T-REx~\cite{trex}. These datasets cover a wide range of scenarios, including open-domain QA (NQ, TriviaQA, MARCO QA), multi-hop reasoning (HotpotQA), and slot filling (T-REx), making them suitable for evaluating knowledge utilization in language models. For all datasets, we adopt Exact Match (EM) and F1 score as our primary evaluation metrics. We use bge-large~\cite{xiao2024c} to retrieve documents from the MS MARCO 2.1~\cite{bajaj2016ms} following \citealp{li2024rag}.

\paragraph{Models and baselines.}
We compare our method across Autoregressive (AR) models and Masked Diffusion Models (MDMs). For AR models, we utilize LLaMA-3.1-8B-Instruct~\cite{grattafiori2024llama} and Qwen2.5-7B-Instruct~\cite{hui2024qwen2}. For MDMs, we evaluate LLaDA-8B-Instruct~\cite{llada} and Dream-v0-Instruct-7B~\cite{dream}. We benchmark ARAM against the following decoding strategies, which can be broadly categorized into AR-adapted contrastive methods and MDM-specific generation mechanisms:
\begin{itemize}
    \item \textbf{Standard RAG:} The conventional generation method for MDMs utilizing the logit conditioned on the retrieved context $\ell_{\text{cond}}(x)$.
    \item \textbf{AR-Based Baselines:} We adapt AR-based methods to the diffusion generation process. These include CAD~\cite{shi2024trusting}, COIECD~\cite{yuan2024discerning}, and \textsc{AdaCAD}~\cite{wang2025adacad}.
    \item \textbf{MDM-Specific Baselines:} We compare against guidance mechanisms tailored for discrete diffusion. These include SPREAD~\cite{yu2026unlocking}, which guides token selection to maintain semantic coherence, and A-CFG~\cite{li2025adaptive}, which dynamically re-masks low-confidence tokens to construct a localized unconditional input for more effective guidance.
\end{itemize}

\subsection{Benchmark Results}
Table~\ref{tab:main_results} presents the overall EM and F1 scores across the five QA benchmarks. The results demonstrate the strong capability of our proposed method in enhancing MDM-based RAG. Compared to the standard MDM-based RAG (which uses a fixed guidance scale $\lambda = 1$), our adaptive guidance improves overall EM and F1 trends.

Applying AR-specific dynamic decoding methods (CAD, COIECD, and \textsc{AdaCAD}) to MDMs results in suboptimal performance. Notably, CAD exhibits consistent degradation in our experiments, suggesting that uniform contrastive guidance can over-correct in MDM-RAG settings. While \textsc{AdaCAD} dynamically adjusts the guidance scale across generation steps, it applies a uniform scale across tokens. This limits performance in MDMs, where token-wise uncertainty varies across positions and denoising steps.

We also compare our method against SPREAD and A-CFG tailored for MDMs. While SPREAD and A-CFG generally improve upon static RAG by intervening in the token unmasking process, our method achieves higher overall performance across most benchmarks. Although SPREAD shows higher scores in the T-REx slot-filling task on Dream, our method outperforms the baselines in open-domain and multi-hop reasoning tasks. 
These results show that calibrating logit-space guidance via SNR provides a strong alternative to re-masking strategies, achieving better overall performance.

\begin{table}[t]
\centering
\caption{Categorization rates of retrieval-prior interaction outcomes.}

\label{tab:prior_behavior}
\resizebox{\columnwidth}{!}{
\begin{tabular}{lcccc}
\toprule
\textbf{Method} & \textbf{Positive ($\uparrow$)} & \textbf{Negative ($\downarrow$)} & \textbf{Cons. Correct} & \textbf{Cons. Wrong} \\
\midrule
\multicolumn{5}{c}{\textit{LLaDA}} \\
\midrule
RAG  & 0.135 & 0.051 & 0.040 & 0.774 \\
\rowcolor{cyan!10}
\textbf{Ours} & \textbf{0.234} & \textbf{0.022} & \textbf{0.069} & \textbf{0.675} \\
\midrule
\multicolumn{5}{c}{\textit{Dream}} \\
\midrule
RAG & 0.259 & 0.049 & 0.100 & 0.592 \\
\rowcolor{cyan!10}
\textbf{Ours} & \textbf{0.266} & \textbf{0.038} & \textbf{0.111} & \textbf{0.585} \\
\bottomrule
\end{tabular}
}
\end{table}

\begin{table}[t]
\centering
\caption{Performance comparison across subsets stratified by retrieval context quality.}
\label{tab:noise_robustness}
\resizebox{\columnwidth}{!}{
\begin{tabular}{lcccccc}
\toprule
\multirow{2}{*}{\textbf{Method}} 
& \multicolumn{2}{c}{\textbf{Gold}} 
& \multicolumn{2}{c}{\textbf{Non-Answering}} 
& \multicolumn{2}{c}{\textbf{Irrelevant}} \\
\cmidrule(lr){2-3} \cmidrule(lr){4-5} \cmidrule(lr){6-7}
& LLaDA & Dream & LLaDA & Dream & LLaDA & Dream \\
\midrule
RAG & 24.2 & 49.7 & 0.77 & 1.54 & 0.00 & 0.00 \\
\rowcolor{cyan!10}
\textbf{Ours} & \textbf{41.7} & \textbf{52.0} & \textbf{1.92} & \textbf{1.92} & \textbf{0.00} & \textbf{0.00} \\
\bottomrule
\end{tabular}
}
\vspace{-0.3cm}
\end{table}

\subsection{Analysis of Retrieval-Prior Conflict}
\label{sec:analysis}

To validate how ARAM resolves the knowledge conflict between the parametric prior and retrieved context, we analyze both model behavior and retrieval context quality. Detailed settings are provided in Appendix~\ref{sec:implementation_detail}.

First, we analyze the interaction between retrieval context and prior knowledge by comparing retrieval-based generation with the retrieval-free baseline. We categorize the outcomes into four cases: \textit{Positive} (baseline incorrect $\rightarrow$ retrieval-based correct), \textit{Negative} (baseline correct $\rightarrow$ retrieval-based incorrect), \textit{Consistently Correct}, and \textit{Consistently Wrong}. As shown in Table~\ref{tab:prior_behavior}, ARAM increases the \textit{Positive} rate while reducing the \textit{Negative} rate for both models compared to static RAG. This indicates that ARAM strengthens guidance when retrieval provides useful corrective evidence while suppressing it when the retrieved context conflicts with the parametric prior.

\begin{figure}[t]
    \centering
    \begin{subfigure}[b]{0.48\textwidth}
        \centering
        \includegraphics[width=0.9\textwidth]{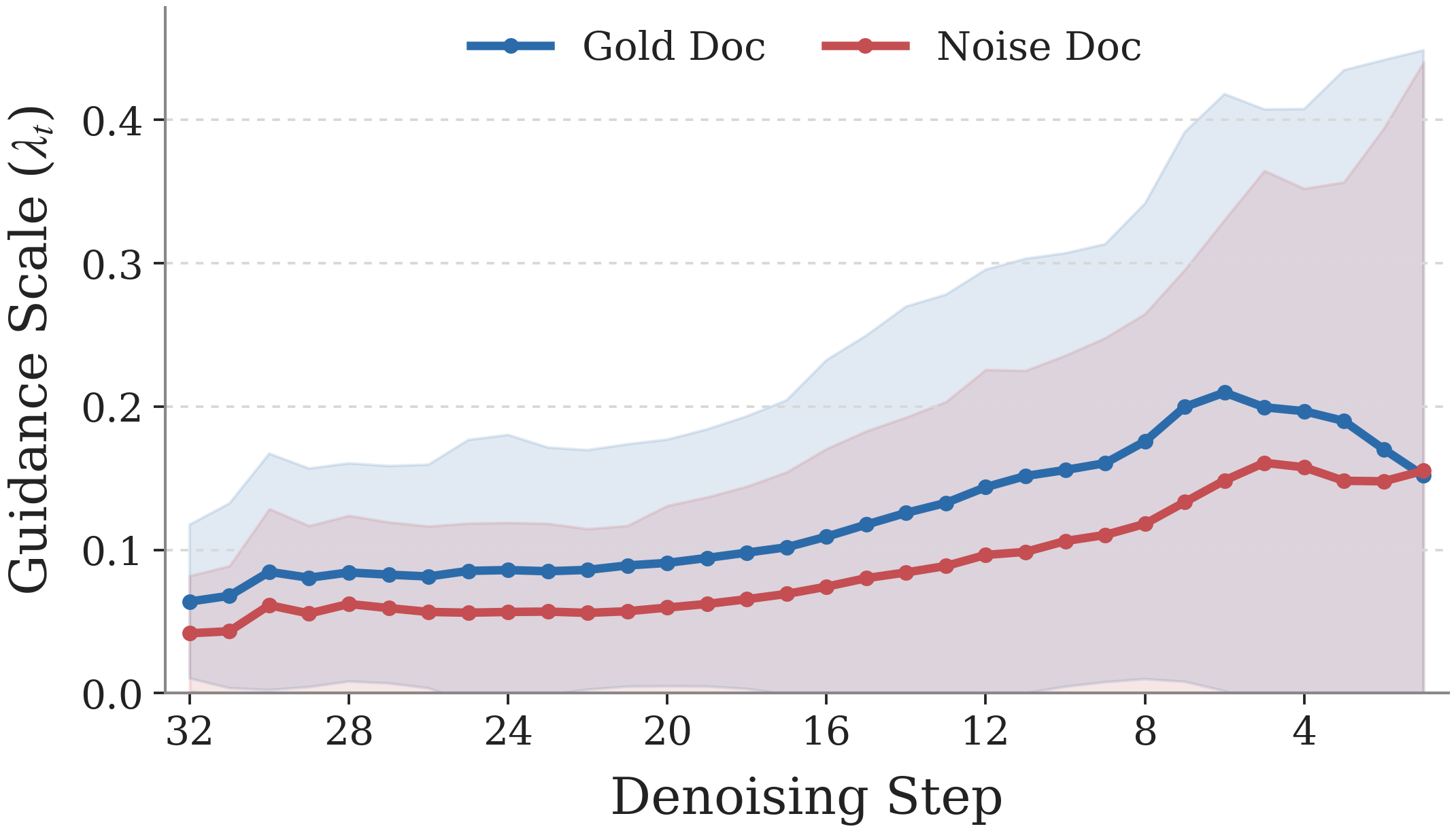}
        \caption{Average guidance trajectory.}
        \vspace{0.2cm}
        \label{fig:adaptive_lambda_a}
    \end{subfigure}
    \hfill
    \begin{subfigure}[b]{0.4\textwidth}
        \centering
        \includegraphics[width=0.9\textwidth]{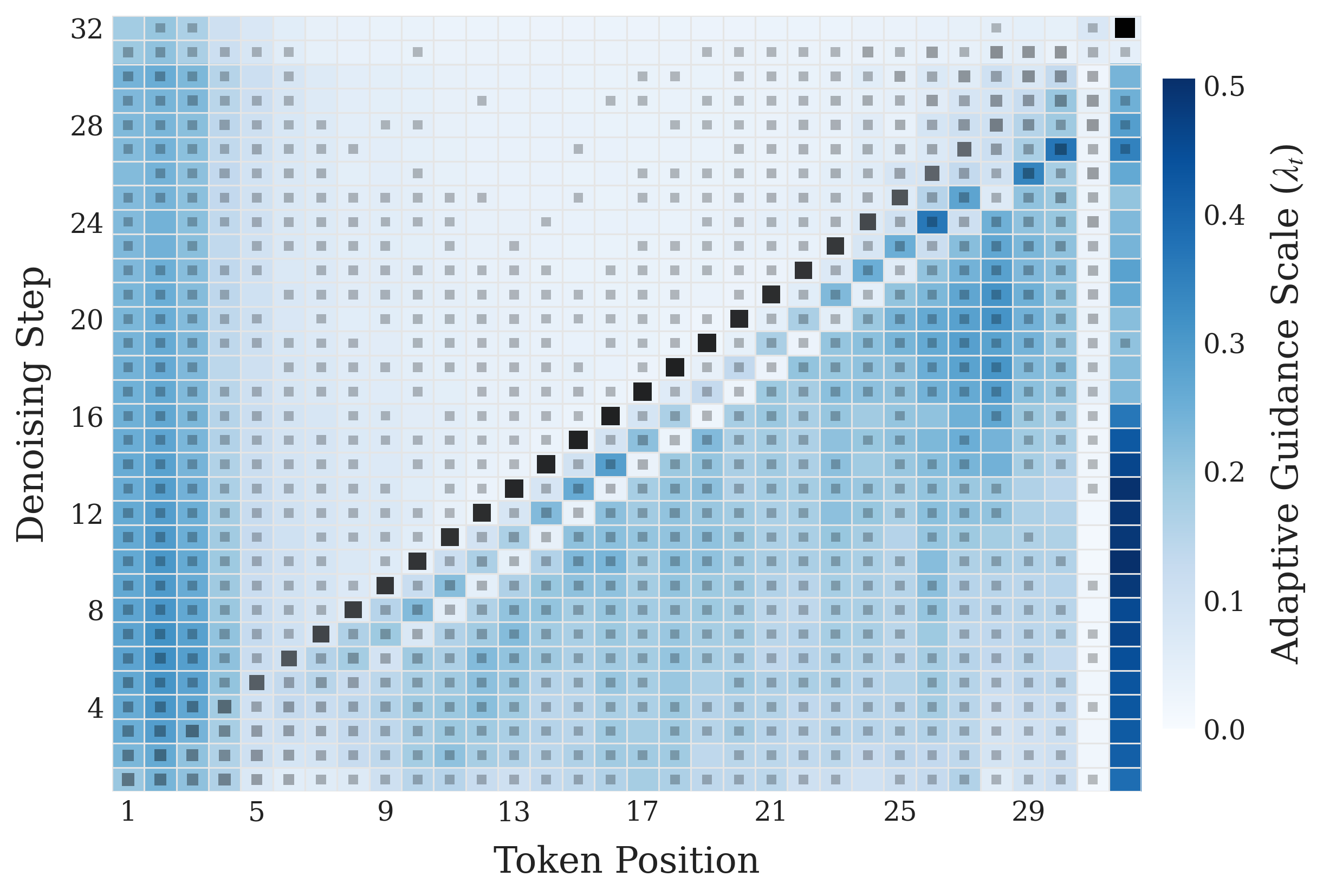}
        \caption{Token-level guidance heatmap.}
        \vspace{-0.1cm}
        \label{fig:adaptive_lambda_b}
    \end{subfigure}
    \caption{\textbf{Dynamics of adaptive guidance}. \textbf{(a)} Average guidance scale ($\lambda_t$) per step. ARAM maintains high guidance for gold documents but suppresses $\lambda_t$ for noisy ones. \textbf{(b)} Token-wise guidance heatmap. Black squares mark unmasking positions, demonstrating a right-to-left generation tendency. ARAM assigns larger guidance to positions that align with the unmasking pattern.}
    \label{fig:analysis_main}
    \vspace{-0.5cm}
\end{figure}

Second, we stratify the evaluation samples by retrieval context quality. We first define \textit{Gold} as cases where the retrieved context contains the answer. For the remaining samples (equivalent to \textit{Noise Doc} in Sec.~\ref{subsec:snr_proxy}), we further distinguish \textit{Non-Answering} as cases where the retrieved context is related to the query but does not provide sufficient evidence for the correct answer, and \textit{Irrelevant} as cases where the retrieved context is off-topic. Table~\ref{tab:noise_robustness} shows that ARAM improves over static RAG in both the \textit{Gold} and \textit{Non-Answering} settings. On \textit{Gold} samples, ARAM more effectively incorporates valid external evidence, mitigating the under-guidance problem of static RAG. On the \textit{Non-Answering} subset, ARAM avoids over-amplifying unreliable contextual signals. On the \textit{Irrelevant} subset, ARAM does not improve over static RAG, indicating that adaptive guidance cannot recover performance when the retrieved context is entirely uninformative.

Taken together, these results show that adaptive SNR-based guidance better resolves the dilemma of static guidance under knowledge conflict. ARAM reduces harmful retrieval-induced changes while preserving beneficial corrections and remains effective when the retrieved context is semantically related but non-supportive.

\begin{table}[t]
\centering
\small
\caption{Ablation study on $\lambda_{\max}$ and $\beta$. Bold denotes the best score and underlined represents second-best.}
\label{tab:ablation}
\begin{tabular}{cc|cc|cc}
\toprule
\multirow{2}{*}{$\lambda_{\max}$} & \multirow{2}{*}{$\beta$} & \multicolumn{2}{c|}{\textbf{LLaDA}} & \multicolumn{2}{c}{\textbf{Dream}} \\

 & & \textbf{EM} & \textbf{F1} & \textbf{EM} & \textbf{F1} \\
\midrule
 1.0 &\multirow{3}{*}{1.0} & \cellcolor{cyan!10}\textbf{24.9} & \cellcolor{cyan!10}\textbf{36.1} & \cellcolor{cyan!10}\textbf{33.9} & \cellcolor{cyan!10}\textbf{43.6} \\
 2.0& & \underline{13.9} & \underline{27.3} & \underline{31.4} & \underline{42.4}\\
 3.0& & 9.1 & 22.8 & 26.3 & 38.9 \\
 
\midrule
\multirow{6}{*}{1.0} & 0.1  & \cellcolor{cyan!10}\textbf{30.4} & \cellcolor{cyan!10}\textbf{39.3} & 33.5 & 42.0 \\
&0.3   & \underline{28.4} & \underline{38.7} & \textbf{34.3} & 43.4 \\
&0.5   & 26.9 & 37.7 & \cellcolor{cyan!10}\underline{34.2} & \cellcolor{cyan!10}\underline{43.5} \\
&0.7  & 26.1 & 37.0 & 34.0 & \underline{43.5} \\
&1.0   & 24.9 & 36.1 & 33.9 & \textbf{43.6} \\
&2.0   & 22.3 & 34.1 & 33.6 & 43.3 \\
\bottomrule
\end{tabular}
\vspace{-0.2cm}
\end{table}

\subsection{Dynamics of Adaptive Guidance}
\label{sec:dynamics}
To analyze the behavior of the proposed adaptive guidance, we monitor the step-wise and token-wise guidance scales during the denoising process, as shown in Fig.~\ref{fig:analysis_main}. We sampled 200 instances from the NQ dataset for both \textit{Gold Doc} (documents containing the answer) and \textit{Noise Doc} (documents not containing the answer) settings using the LLaDA model. At each timestep $t$, we recorded the token-wise average $\lambda_t$ values. 

As illustrated in Fig.~\ref{fig:adaptive_lambda_a}, the step-wise average guidance trajectory aligns with the SNR formulation. When provided with a \textit{Gold Doc}, the model maintains a larger guidance scale. In contrast, when confronted with a \textit{Noise Doc}, the guidance scale is decreased, effectively mitigating the risk of the model hallucinating based on irrelevant context.

Furthermore, we visualize the token-level guidance heatmap to examine the spatial distribution of $\lambda_t$. In Fig.~\ref{fig:adaptive_lambda_b}, LLaDA exhibits a right-to-left unmasking pattern. Specifically, the highest guidance values cluster adjacent to recently unmasked tokens. Since newly generated tokens provide strong local context for predicting nearby masked tokens, this spatial distribution suggests that ARAM allocates stronger guidance to positions that remain uncertain near recently resolved tokens.

\subsection{Ablation Study}
\paragraph{Effect of guidance hyperparameters.}
We investigate the impact of the maximum guidance scale $\lambda_{\max}$ and the sensitivity parameter $\beta$ on generation quality. As shown in Table~\ref{tab:ablation}, varying $\beta$ while fixing $\lambda_{\max}=1.0$ reveals that the two models exhibit different sensitivities to the context signal. Specifically, Dream achieves balanced performance at $\beta=0.5$, while LLaDA achieves optimal results at $\beta=0.1$.

\begin{table}[t]
\centering
\small
\caption{Comparison of noise proxies. Average EM and F1 scores are reported.}
\label{tab:noise_proxy}
\begin{tabular}{l|cc|cc}
\toprule
\multirow{2}{*}{\textbf{Noise Proxy}} & \multicolumn{2}{c|}{\textbf{LLaDA}} & \multicolumn{2}{c}{\textbf{Dream}} \\
 & \textbf{EM} & \textbf{F1} & \textbf{EM} & \textbf{F1} \\
\midrule
$\mathrm{Var}_{p_{\text{prior}}}(s)$ & 20.3 & 32.5 & 33.5 & 43.2 \\
\rowcolor{cyan!10}
$\mathcal{H}(p_{\text{cond}})$ & \textbf{24.9} & \textbf{36.1} & \textbf{33.9} & \textbf{43.6} \\
\bottomrule
\end{tabular}
\vspace{-0.2cm}
\end{table}

\begin{table}[t]
\centering
\small
\caption{Performance on LLaDA 1.5.}
\vspace{-0.2cm}
\label{tab:llada15}
\begin{tabular}{l|cc}
\toprule
\textbf{Method} & \textbf{EM} & \textbf{F1} \\
\midrule
LLaDA 1.5 & 15.8 & 22.0 \\
LLaDA 1.5 + RAG & 17.6 & 30.6 \\
\rowcolor{cyan!10}
LLaDA 1.5 + Ours & \textbf{24.2} & \textbf{35.9} \\
\bottomrule
\end{tabular}
\vspace{-0.16cm}
\end{table}

\paragraph{Choice of noise proxy.}
To validate the use of conditional entropy $\mathcal{H}(p_{\text{cond}})$ as a noise proxy, we compare it against the score variance $\mathrm{Var}_{p_{\text{prior}}}(s)$ derived in \eqref{eq:optimal_lambda}. For a controlled comparison, we set $\lambda_{\max}=1.0$ and $\beta=1.0$ for both proxies. As shown in Table~\ref{tab:noise_proxy}, the entropy proxy yields higher generation quality for both models. This performance gap stems from the behavior of noise proxies during the unmasking process (Fig.~\ref{fig:variance_unstable}), demonstrating that the conditional entropy provides a more stable measure of uncertainty, resulting in robust guidance. Additional ablation studies on guidance scale design are provided in Appendix~\ref{sec:additional_ablation}.

\paragraph{Scalability across  model configurations.}
To verify the generalizability of ARAM, we evaluate the framework on a more recent architecture, LLaDA 1.5~\cite{zhu2025llada}. As shown in Table~\ref{tab:llada15}, ARAM improves the EM and F1 scores over the standard RAG baseline using $\lambda_{\max}=1.0, \beta=1.0$. This outcome indicates that our SNR-based adaptive guidance mechanism functions consistently across different model variants.

\section{Conclusion}

In this work, we introduce Adaptive Retrieval-Augmented Masked Diffusion (ARAM), a decoding framework designed to address knowledge conflict in MDM-based RAG. We propose an adaptive guidance rule based on a Signal-to-Noise Ratio, which suppresses guidance when the retrieved context is conflicting or uninformative. Extensive experiments across QA benchmarks demonstrate that ARAM improves robustness under retrieval-prior conflict and demonstrates strong improvements over competitive MDM-based RAG baselines. 

\section*{Limitations}
While our findings demonstrate that ARAM effectively mitigates knowledge conflict in MDM-based RAG, we acknowledge several limitations. Our empirical claims are scoped to English-language, knowledge-intensive QA tasks using open-source MDMs. It remains an open question whether our method translates effectively to open-ended generation tasks or other decoding strategies. From a practical perspective, computing the guidance scale across the entire vocabulary at every denoising step introduces a computational overhead during inference. As quantified in Appendix~\ref{sec:computational_cost}, ARAM incurs additional inference latency compared to standard CFG due to these intermediate calculations. Due to this overhead, it remains an important consideration for deploying ARAM in highly latency-sensitive production environments.

\bibliography{references}

\newpage
\appendix

\section{Proofs and Derivations}
\label{sec:proof}

\subsection{Proof of Theorem~\ref{theorem:1}}
The Donsker-Varadhan (DV) variational representation~\cite{donsker1983asymptotic} of the Kullback-Leibler divergence for two distributions $P$ and $Q$ states for the function class $f:\mathcal{X} \to \mathbb{R}$ that:
\begin{equation*}
D_{\mathrm{KL}}(P \| Q) = \sup_{f} \left( \mathbb{E}_{P}[f(x)] - \log \mathbb{E}_{Q}[\exp{(f(x))}] \right).
\end{equation*}
We set $P = p_{\text{cond}}$, $Q = p_{\text{prior}}$, and $f(x) = \lambda s(x;\mathcal{C}) = \lambda \log \frac{p_{\text{cond}}(x)}{p_{\text{prior}}(x)}$ for $\lambda \in \mathbb{R}$. Then, substituting these into the DV representation yields:
\begin{equation*}
D_{\mathrm{KL}} \ge \mathbb{E}_{p_{\text{cond}}}[\lambda s(x;\mathcal{C})] - \log \mathbb{E}_{p_{\text{prior}}}[\exp({\lambda s(x;\mathcal{C}))}].
\end{equation*}
By definition, $\mathrm{IG}_t = D_{\mathrm{KL}}(p_{\text{cond}} \| p_{\text{prior}})$ and $Z_\lambda = \mathbb{E}_{p_{\text{prior}}}[\exp(\lambda s(x;\mathcal{C}))]$. Therefore, we obtain the lower bound:
\begin{equation*}
\mathrm{IG}_t \ge \lambda \mathbb{E}_{p_{\text{cond}}}[s(x;\mathcal{C})] - \log Z_\lambda =: \mathcal{L}(\lambda).
\end{equation*}
\qed

\subsection{Proof of Corollary~\ref{coro:dv_max}}
To evaluate $\mathcal{L}(\lambda)$ at $\lambda=1$, we first expand the partition term $Z_1$:
\begin{align*}
Z_1 &= \mathbb{E}_{p_{\text{prior}}}[\exp(s(x;\mathcal{C}))]\\ 
&= \sum_{x \in \mathcal{V}} p_{\text{prior}}(x) \exp\left( \log \frac{p_{\text{cond}}(x)}{p_{\text{prior}}(x)} \right) \\
&= \sum_{x \in \mathcal{V}} p_{\text{cond}}(x) \\
&= 1.
\end{align*}
Consequently, the log-partition function is $\log Z_1 = 0$. Substituting this into $\mathcal{L}(1)$ gives:
\begin{align*}
\mathcal{L}(1) &= 1 \cdot \mathbb{E}_{p_{\text{cond}}}[s(x;\mathcal{C})] - \log(1) \\
&= \mathbb{E}_{p_{\text{cond}}}[s(x;\mathcal{C})]\\
&=\sum_{x\in\mathcal{V}}p_{\text{cond}} \log \frac{p_{\text{cond}}}{p_{\text{prior}}} \\
&=D_{\mathrm{KL}}(p_{\text{cond}} \| p_{\text{prior}}).
\end{align*}
Since $\mathcal{L}(\lambda)$ is bounded above by $D_{\mathrm{KL}}(p_{\text{cond}} \| p_{\text{prior}})$ according to Theorem~\ref{theorem:1}, $\lambda=1$ is a global maximizer. 
\qed


\subsection{Derivation of \eqref{eq:optimal_lambda}}
\label{}
We analyze the local geometry of $\mathcal{L}(\lambda)$ around the parametric anchor $\lambda=0$. The objective function is:
\begin{equation*}
\mathcal{L}(\lambda) = \lambda \mathbb{E}_{p_{\text{cond}}}[s(x;\mathcal{C})] - \log Z_\lambda,    
\end{equation*}
where the partition function $Z_\lambda$ is:
\begin{align*}
 Z_\lambda &= \mathbb{E}_{p_{\text{prior}}}[\exp(\lambda s(x;\mathcal{C}))]   \\
 &= \sum_{x \in \mathcal{V}}p_{\text{prior}} \exp(\lambda s(x;\mathcal{C})).
\end{align*}
and $s(x;\mathcal{C}) = \log \frac{p_{\text{cond}}(x)}{p_{\text{prior}}(x)}.$

The first derivative with respect to $\lambda$ is:
\begin{equation*}
\mathcal{L}'(\lambda) = \mathbb{E}_{p_{\text{cond}}}[s(x;\mathcal{C})] - \frac{\partial}{\partial \lambda} \log Z_\lambda.
\end{equation*}
Using the property of the log-partition function, its derivative is the expectation under the tilted distribution $p_\lambda(x):= \frac{p_{\text{prior}}(x) \exp(\lambda s(x;\mathcal{C}))}{Z_\lambda}$:
\begin{align*}
\frac{\partial}{\partial \lambda} \log Z_\lambda &= \frac{1}{Z_\lambda} \sum_{x \in \mathcal{V}} p_{\text{prior}}(x) s(x;\mathcal{C}) \exp(\lambda s(x;\mathcal{C}))\\
&=\sum_{x \in \mathcal{V}} p_\lambda(x) s(x;\mathcal{C})\\
&= \mathbb{E}_{p_\lambda}[s(x;\mathcal{C})].
\end{align*}
Thus, $\mathcal{L}'(\lambda) = \mathbb{E}_{p_{\text{cond}}}[s(x;\mathcal{C})] - \mathbb{E}_{p_\lambda}[s(x;\mathcal{C})].$ At $\lambda=0$, $p_0(x) = p_{\text{prior}}(x)$, which yields:
\begin{align*}
\mathcal{L}'(0) = \mathbb{E}_{p_{\text{cond}}}[s(x;\mathcal{C})] - \mathbb{E}_{p_{\text{prior}}}[s(x;\mathcal{C})].
\end{align*}
Using the first derivative, the second derivative with respect to $\lambda$ is:
\begin{align*}
\mathcal{L}''(\lambda) &= \frac{\partial}{\partial \lambda} \mathcal{L'} \\
&= \frac{\partial}{\partial \lambda} \left(\mathbb{E}_{p_{\text{cond}}}[s(x;\mathcal{C})] - \mathbb{E}_{p_\lambda}[s(x;\mathcal{C})]\right) \\
&= -\frac{\partial}{\partial \lambda}\mathbb{E}_{p_\lambda}[s(x;\mathcal{C})] \\
&= -\sum_{x \in \mathcal{V}}s(x;\mathcal{C})\frac{\partial p_\lambda(x)}{\partial \lambda}.
\end{align*}
To get the derivative of $p_\lambda(x)$, we first calculate derivative of log-probability as:
\begin{align*}
\frac{\partial}{\partial \lambda} \log p_\lambda(x) &= \frac{\partial}{\partial \lambda} \Big(\log p_{\text{prior}}(x) +\lambda s(x;\mathcal{C}) - \log Z_\lambda \Big)\\
&= s(x;\mathcal{C}) - \frac{\partial}{\partial \lambda} \log Z_\lambda \\
&= s(x;\mathcal{C}) - \mathbb{E}_{p_\lambda} [s(x;\mathcal{C})].
\end{align*}
Then, we obtain 
\begin{align*}
 \frac{\partial}{\partial \lambda}p_\lambda(x) &= p_\lambda(x) \frac{\partial}{\partial \lambda} \log p_\lambda(x)\\
 &= p_\lambda(x)\Big(s(x;\mathcal{C})-\mathbb{E}_{p_\lambda}[s(x;\mathcal{C})]\Big).   
\end{align*}
Therefore, we calculate 
\begin{align*}
\mathcal{L''}(\lambda) &= -\sum_{x \in \mathcal{V}} s(x;\mathcal{C}) p_\lambda(x) \Big( s(x;\mathcal{C})-\mathbb{E}_{p_\lambda}[s(x;\mathcal{C})]\Big) \\
&= - \Big( \mathbb{E}_{p_\lambda}[s(x;\mathcal{C})^2] - (\mathbb{E}_{p_\lambda}[s(x;\mathcal{C})])^2 \Big) \\
&= -\mathrm{Var}_{p_\lambda}(s(x;\mathcal{C})).
\end{align*}
Evaluating at $\lambda=0$ gives:
\begin{align*}
\mathcal{L}''(0) = - \mathrm{Var}_{p_{\text{prior}}}(s(x;\mathcal{C}))
\end{align*}
Since $Z_0 = 1$, we have $\mathcal{L}(0) = 0$. The second-order Taylor expansion of $\mathcal{L}(\lambda)$ around $\lambda=0$ is:
\begin{align*}
\mathcal{L}(\lambda) &\approx \mathcal{L}(0) + \lambda \mathcal{L}'(0) + \frac{\lambda^2}{2} \mathcal{L}''(0) \\
&= \lambda \left( \mathbb{E}_{p_{\text{cond}}}[s] - \mathbb{E}_{p_{\text{prior}}}[s] \right) - \frac{\lambda^2}{2} \mathrm{Var}_{p_{\text{prior}}}(s).
\end{align*}
To find the local maximum of this concave quadratic approximation, we set the derivative with respect to $\lambda$ to zero:
\begin{align*}
\mathbb{E}_{p_{\text{cond}}}[s] - \mathbb{E}_{p_{\text{prior}}}[s] - \lambda \mathrm{Var}_{p_{\text{prior}}}(s) = 0.
\end{align*}
Solving for $\lambda$ yields the optimal scale $\lambda^*$ formulated as the Signal-to-Noise Ratio:
\begin{align*}
\lambda^* = \frac{\mathbb{E}_{p_{\text{cond}}}[s]-\mathbb{E}_{p_{\text{prior}}}[s]}{\mathrm{Var}_{p_{\text{prior}}}(s)}.
\end{align*}

\section{Implementation Details}
\label{sec:implementation_detail}

\paragraph{Hyperparameters.} In the retrieval component, we fetch the top-3 relevant documents per query. All benchmark evaluations are conducted on a randomly sampled subset of 1,000 instances per dataset. For all baseline methods, we adopt the default hyperparameters specified in their original papers. For the autoregressive baselines adapted to MDMs, we set the guidance scale to $\lambda=2$ for CAD, and use $\alpha=1$ with a threshold of $0.25$ for COIECD. In AdaCAD, we substitute the static $\lambda$ with the token-level Jensen-Shannon Divergence (JSD). For MDM-specific baselines, SPREAD modifies the re-masking strategy without altering the pre-softmax logits. For A-CFG, we use a re-masking proportion of $\rho=0.7$ and a guidance scale of $\lambda=1.5$ (corresponding to $w=0.5$ in the original formulation). For the proposed ARAM, we calculate $\lambda_t$ at each denoising step, setting the sensitivity parameter to $\beta=0.1$ and the maximum guidance scale to $\lambda_{\max}=1.0$ for both LLaDA and Dream.

\paragraph{Prompt and Decoding Settings.} We follow the decoding configuration and prompt design of Attention Floating~\cite{dai2026revealing}. We utilize a temperature of 0, nucleus sampling with $p=0.9$, a maximum generation length of 32 tokens, and 32 denoising steps. For the unmasking strategy, we adopt the default policies of the respective backbone models: low-confidence for LLaDA and entropy for Dream. We employ a zero-shot instruction prompt for LLaDA (Figure~\ref{fig:prompt_llada}) and a few-shot prompt with in-context examples for Dream and other AR baselines (Figure~\ref{fig:prompt_dream}). When retrieved documents are available, they are concatenated sequentially with index labels (e.g., ``Passage 1: [Text]''). In the no-retrieval setting, the context field is populated with the string ``No relevant context available.''

\paragraph{Experiment Details for Robustness Analysis.} 
For the analysis in Sec.~\ref{sec:analysis}, we utilize the outputs from the main experiments: the retrieval-free baseline, static RAG, and ARAM. For each backbone model, we align predictions across methods on the NQ benchmark and determine correctness using the exact-match criterion.

For the retrieval-prior interaction analysis (Table~\ref{tab:prior_behavior}), we compare each retrieval-based method against the retrieval-free baseline on an instance-by-instance basis. Each sample is assigned to one of four categories: \textit{Positive} if the no-retrieval prediction is incorrect but the retrieval-based method is correct, \textit{Negative} if the no-retrieval prediction is correct but the retrieval-based method is incorrect, \textit{Consistently Correct} if both are correct, and \textit{Consistently Wrong} if both are incorrect. We report the proportion of instances in each category.

For the analysis of the retrieval context quality (Table~\ref{tab:noise_robustness}), we stratify samples according to the quality of the retrieved context. We first define a sample as \textit{Gold} if at least one retrieved document contains the answer. This is determined by normalized string matching, including lowercasing, article removal, and punctuation stripping, between the reference answer set and the retrieved documents. For the remaining non-gold (equivalent to \textit{Noise Doc} in Sec.~\ref{sec:dynamics}) samples, we further distinguish between \textit{Non-Answering} and \textit{Irrelevant} cases using an LLM-based annotation procedure that utilizes the prompt template detailed in Figure~\ref{fig:prompt_llm_eval}. Given the question, reference answer, and retrieved documents, the LLM is prompted to output one of two labels: \textit{irrelevant}, if the documents are off-topic and provide no useful evidence for answering the question, or \textit{Non-Answering}, if the documents are topically related but fail to provide sufficient support for the correct answer. For stable annotation, we use GPT-4.1-mini with deterministic decoding.

\section{Additional Ablation Studies}
\label{sec:additional_ablation}
\begin{table}[t]
\centering
\small
\caption{Comparison of noise proxies. Average EM and F1 scores are reported.}
\label{tab:noise_proxy_more}
\begin{tabular}{l|cc|cc}
\toprule
\multirow{2}{*}{\textbf{Noise Proxy}} & \multicolumn{2}{c|}{\textbf{LLaDA}} & \multicolumn{2}{c}{\textbf{Dream}} \\
 & \textbf{EM} & \textbf{F1} & \textbf{EM} & \textbf{F1} \\
\midrule
$\mathrm{Var}_{p_{\text{prior}}}(s)$ & 20.3 & 32.5 & 33.5 & 43.2 \\
$\mathcal{H}(p_{\text{prior}})$ & 20.5 & 32.4 & 33.6 & 42.8 \\
$|\mathcal{H}(p_{\text{prior}}) - \mathcal{H}(p_{\text{cond}})|$ & 23.7 & 35.3 & 33.4 & 42.8 \\
\rowcolor{cyan!10}
$\mathcal{H}(p_{\text{cond}})$ & \textbf{24.9} & \textbf{36.1} & \textbf{33.9} & \textbf{43.6} \\
\bottomrule
\end{tabular}
\end{table}

\begin{table}[t]
\centering
\small
\caption{Comparison of stability function. Average EM and F1 scores are reported.}
\label{tab:stability}
\begin{tabular}{l|cc|cc}
\toprule
\multirow{2}{*}{\textbf{Signal Proxy}} & \multicolumn{2}{c|}{\textbf{LLaDA}} & \multicolumn{2}{c}{\textbf{Dream}} \\
 & \textbf{EM} & \textbf{F1} & \textbf{EM} & \textbf{F1} \\
\midrule
- (Raw SNR) & 21.5 & 33.5 & 33.5 & 43.4 \\
\rowcolor{cyan!10}
Tanh & \textbf{24.9} & \textbf{36.1} & \textbf{33.9} & \textbf{43.6} \\
\bottomrule
\end{tabular}
\end{table}

\paragraph{Extended Comparison of Noise Proxies.}
To further investigate the formulation of the noise proxy, we compare the conditional entropy $\mathcal{H}(p_{\text{cond}})$ against three alternative metrics: score variance $\mathrm{Var}_{p_{\text{prior}}}(s)$, prior entropy $\mathcal{H}(p_{\text{prior}})$, and the absolute entropy difference $|\mathcal{H}(p_{\text{prior}}) - \mathcal{H}(p_{\text{cond}})|$. As shown in Table~\ref{tab:noise_proxy_more}, $\mathcal{H}(p_{\text{cond}})$ yields the highest EM and F1 scores across both Dream and LLaDA models. The prior entropy $\mathcal{H}(p_{\text{prior}})$ omits the uncertainty introduced by the external context, leading to lower guidance accuracy. The absolute difference formulation also underperforms compared to direct conditional entropy. This indicates that $\mathcal{H}(p_{\text{cond}})$ alone provides a sufficient and stable measure for the noise introduced by the retrieved context.

\paragraph{Effect of the Stability Function.}
Table~\ref{tab:stability} presents the ablation on the stability function used to bound the adaptive guidance scale. We compare the raw SNR scaling against the Tanh-scaled formulation. The results demonstrate that applying the Tanh function improves generation performance for both models. This improvement is attributed to the bounding property of Tanh, which prevents numerical explosion of $\lambda_t$ when the denominator (noise) approaches zero during the unmasking steps. Bounding the guidance scale ensures stable generation without over-amplifying the contextual logits.

\paragraph{Computational Cost Analysis.}
\label{sec:computational_cost}

We measure inference latency to analyze the computational overhead introduced by retrieval and adaptive guidance. Table~\ref{tab:comp_cost} reports the average execution time over three runs for both the LLaDA and Dream backbones. The latency gap between No-RAG and static RAG mainly comes from the longer input sequence induced by concatenating retrieved context. In contrast, methods based on classifier-free guidance, including standard CFG and ARAM, require both conditional and unconditional forward passes at each step (2 NFE), leading to higher overall latency. Compared with standard CFG, ARAM introduces additional overhead from the computation of the adaptive guidance scale, including the entropy term and the Signal-to-Noise Ratio.

\begin{table}[t]
\centering
\small
\caption{Average inference time per query (seconds).}
\label{tab:comp_cost}
\begin{tabular}{l|c|cc}
\toprule
\textbf{Method} & \textbf{NFE / Step} & \textbf{LLaDA} & \textbf{Dream} \\
\midrule
No-RAG & 1 & 0.93 & 0.94 \\
RAG & 1 & 1.55 & 1.56 \\
CFG & 2 & 2.02 & 3.06 \\
\rowcolor{cyan!10}
ARAM (Ours) & 2 & 2.07 & 3.29 \\
\bottomrule
\end{tabular}
\end{table}

\begin{figure}[!h]
    \centering
    \begin{tcolorbox}[colback=gray!5!white, colframe=gray!75!black, fonttitle=\bfseries, title={LLM Annotation Prompt}]
    \small
    You are labeling retrieval context quality for a QA example.\\

    Question:\\
    \{query\}\\

    Reference answers:\\
    \{answers\}\\

    Retrieved passages:\\
    \{passages\}\\

    Task:\\
    Return exactly one label:\\
    - irrelevant: the passages are largely unrelated to the question and do not provide useful evidence.\\
    - related\_non\_answering: the passages are related to the question but do not clearly contain the correct answer.\\

    Rules:\\
    - If the passages are about the same entity/topic/question but fail to answer it clearly, return non answering.\\
    - Use irrelevant only when the passages are genuinely off-topic or retrieval junk.\\
    - Return only the label, with no explanation.
    \end{tcolorbox}
    \vspace{-2mm}
    \caption{Prompt template for LLM-based retrieval context quality annotation.}
    \label{fig:prompt_llm_eval}
\end{figure}

\begin{figure}[!h]
    \centering
    \begin{tcolorbox}[colback=gray!5!white, colframe=gray!75!black, fonttitle=\bfseries, title={Prompt for LLaDA}]
    \small
    Use the following passages as context if they are helpful, but provide a SHORT and DIRECT answer.
    \\
    
    RULES:\\
    - First, check the context passages for the answer\\
    - If the context contains the answer, use it\\
    - If not, use your general knowledge\\
    - Answer must be 1-10 words maximum\\
    - Do NOT include any explanations or extra text\\
    - Do NOT repeat or summarize the passages\\

    Context: \\
    \{context\}\\

    Question: \{query\}\\

    Short Answer:
    \end{tcolorbox}
    \vspace{-2mm}
    \caption{Prompt template used for LLaDA.}
    \label{fig:prompt_llada}
\end{figure}

\begin{figure}[!h]
    \centering
    \begin{tcolorbox}[colback=gray!5!white, colframe=gray!75!black, fonttitle=\bfseries, title={Prompt for Dream}]
    \small
    You are a helpful assistant that answers questions. \\

    INSTRUCTIONS:\\
    1. First, check if the answer is in the provided context passages\\
    2. If the answer is in the context, use it\\
    3. If the context doesn't contain the answer, use your general knowledge\\
    4. Always provide a direct, concise answer (typically 1-10 words)\\
    5. Do NOT include explanations, reasoning, or phrases like "based on" or "according to"\\
    6. Never say "no answer found" - always attempt to answer using available information\\

    Context:\\
    \{context\}\\
    
    Here are examples showing the expected answer format:\\

    Example 1:\\
    Context:\\
    Passage 1: The Eiffel Tower is located in Paris, France. It was completed in 1889.\\
    Question: Where is the Eiffel Tower located?\\
    Answer: Paris, France\\

    Example 2:\\
    Context:\\
    Passage 1: Albert Einstein was born on March 14, 1879 in Ulm, Germany.\\
    Question: When was Albert Einstein born?\\
    Answer: March 14, 1879\\

    Example 3:\\
    Context:\\
    Passage 1: The Great Wall of China stretches over 13,000 miles.\\
    Question: What is the capital of Japan?\\
    Answer: Tokyo\\
    (Note: This answer uses general knowledge since the context doesn't contain it)\\

    ---\\
    Now answer the following question in the same concise format:\\
    Question: \{query\}\\
    Answer:
    \end{tcolorbox}
    \vspace{-2mm}
    \caption{Few-shot prompt template used for the Dream and other AR backbones.}
    \label{fig:prompt_dream}
\end{figure}

\end{document}